\documentclass{article}

\usepackage{microtype}
\usepackage{graphicx}
\usepackage{booktabs} 

\usepackage{hyperref}

 \usepackage[accepted]{icml2025}
 
\usepackage{amsmath}
\usepackage{amssymb}
\usepackage{mathtools}
\usepackage{amsthm}
\usepackage{bm}
\usepackage[framemethod=TikZ]{mdframed}
\usepackage[most]{tcolorbox}
\usepackage[ruled,vlined]{algorithm2e}
\usepackage{arydshln}
\usepackage{multirow}
\usepackage{subcaption}

\usepackage[capitalize,noabbrev]{cleveref}

\theoremstyle{plain}
\newtheorem{theorem}{Theorem}[section]
\newtheorem{proposition}[theorem]{Proposition}

\theoremstyle{definition}

\theoremstyle{remark}

\mdfsetup{
	middlelinecolor =  none,
	middlelinewidth = 1pt,
	backgroundcolor = cyan!5,
	roundcorner = 5pt,
}

\usepackage[textsize=tiny]{todonotes}

\icmltitlerunning{Generative Unordered Flow}

\begin{document}

\twocolumn[
\icmltitle{Generative Unordered Flow for Set-Structured Data Generation}



\icmlsetsymbol{equal}{*}

\begin{icmlauthorlist}
\icmlauthor{Yangming Li}{yyy}
\icmlauthor{Chaoyu Liu}{yyy}
\icmlauthor{Carola-Bibiane Schönlieb}{yyy}
\end{icmlauthorlist}

\icmlaffiliation{yyy}{Department of Applied Mathematics and Theoretical Physics, University of Cambridge, UK. The second author is the joint first author and the corresponding author of this paper}



\vskip 0.3in
]



\printAffiliationsAndNotice{}  

\begin{abstract}
	Flow-based generative models have demonstrated promising performance across a broad spectrum of data modalities (e.g., image and text). However, there are few works exploring their extension to unordered data (e.g., spatial point set), which is not trivial because previous models are mostly designed for vector data that are naturally ordered. In this paper, we present \textit{unordered flow}, a type of flow-based generative model for set-structured data generation. Specifically, we convert unordered data into an appropriate function representation, and learn the probability measure of such representations through function-valued flow matching. For the inverse map from a function representation to unordered data, we propose a method similar to particle filtering, with Langevin dynamics to first warm-up the initial particles and gradient-based search to update them until convergence. We have conducted extensive experiments on multiple real-world datasets, showing that our \textit{unordered flow} model is very effective in generating set-structured data and significantly outperforms previous baselines.
\end{abstract}

\section{Introduction}

	Flow-based generative models~\citep{lipman2022flow} are shown to be equivalent to diffusion models~\citep{ho2020denoising,song2021scorebased} in the common setup~\citep{gao2025diffusionmeetsflow}, and both models have achieved promising results in a wide range of data modalities, including images~\citep{dhariwal2021diffusion}, videos~\citep{ho2022video}, texts~\citep{li2022diffusion}, audio~\citep{guan24b_interspeech}, and tables~\citep{kotelnikov2023tabddpm,jolicoeur2024generating}. However, there are few works exploring their extensions to set-structured data generation, which is essential in many real applications. For example, point cloud generation~\citep{mao2022pu}, earthquake prediction~\citep{ogata1998space}, and generative view synthesis~\citep{mu2025gsd}.
	
	\paragraph{Point process models lack flexibility.} In probability theory, set-structured data are modeled by the point process~\citep{kingman1992poisson}, with an intensity function to quantify how likely an event occurs. Well-known such models include the Poisson process~\citep{kingman1992poisson} that has constant intensities and the Hawkes process~\citep{hawkes1971spectra} that conditions the intensity on observed events. There are also many works~\citep{mei2017neural,zuo2020transformer} that parameterized the intensity function with neural networks (e.g., Transformer~\citep{vaswani2017attention}). Despite their simplicity and interpretability, those statistical methods lack flexibility in parameterization (e.g., intensity function modeling), resulting in a performance bottleneck. Hence, a number of subsequent works~\citep{Shchur2020Intensity-Free,lyu2022a} sought to develop more general frameworks.
	
	\paragraph{Generative models need to be order-insensitive.} One such direction is to apply the expressive deep generative models (e.g., GAN~\citep{goodfellow2020generative}), which indeed have delivered much better performance than the point process models in generating unordered data. However, generative models are commonly designed for vector data that are naturally ordered, so extra adaptations are required before application. For example, \citet{xie2021generative} constructed an energy-based model~\citep{lecun2006tutorial} with the energy function that is permutation-invariant to the point set, and \citet{bilovs2021scalable} developed a variant of normalizing flow~\citep{kingma2018glow} that can model the probability density of unordered data. 
	
	Both flow-based and diffusion-based generative models need similar adaptations, but the studies on this topic are quite few at present. Very recently, \citet{ludke2023add,ludke2024unlocking} proposed to mimic the diffusion process by randomly adding and deleting discrete points. While their method adopted a similar idea of adding noise and denoising, it is built on top of the point process, which involves many discrete operations (e.g., thinning) and differs from diffusion models (that are based on heat equations~\citep{widder1976heat,sohl2015deep}) in principle. More closely related is \citet{chen2023learning}, which generalized the diffusion process from continuous spaces to integer sets, though this in fact assumes that the set-like data has a certain order. It is also worth noting that there are some works~\citep{mu2025gsd} directly applying diffusion models to unordered data, without considering the permutation invariance.
	
	\paragraph{Our method: unordered flow.} In this paper, we aim to address the gap that there are no adapted diffusion-based or flow-based generative models for fully unordered data. To this end, we introduce \textit{unordered flow}, a flow-based generative model that is permutation-invariant to set-structured data. Specifically, we represent unordered data as a continuous function, which can accurately characterize the discrete structure of a set. The probability measure of this function representation resides in the nice L2 space~\cite{kantorovich2014functional}, which is a nice Hilbert space for function-valued flow matching. We will show that the flow matching in the common Euclidean case can be easily adapted to our functional setting. For the inverse transform that maps a function representation back to the set-structured data, we propose a method similar to particle filtering~\citep{djuric2003particle}, which randomly initializes a number of particles, with gradient ascent to update the particles. Before this gradient-based search, we also apply Langevin dynamics~\citep{coffey2012langevin} to warm-up the initial particles, reducing the impact of noise factors.
	
	In summary, our contributions are as follows:
	\begin{itemize}
		\item Conceptually, we accurately represent the unordered data as a proper functional form, paving the way for function-valued generative modeling;
		\item Technically, we present \textit{unordered flow}, the first flow-based generative model that is permutation-invariant to unordered data, filling a gap in the literature;
		\item Empirically, we have conducted extensive experiments to show that our model can generate unordered data well and notably outperforms the baselines.
	\end{itemize}
	\begin{mdframed}[leftmargin=0pt, rightmargin=0pt, innerleftmargin=10pt, innerrightmargin=10pt, skipbelow=0pt]
		\textbf{\emph{Roadmap}}: In the rest of main text, we first review flow matching and formulate the task of interest in Sec.~\ref{sec: whole preliminaries}; Then, we introduce our method: \textit{unordered flow}, in Sec.~\ref{sec: whole method} and discuss related work in Sec~\ref{sec: related work}; Finally, we present the experiment results in Sec.~\ref{sec: whole experiment}.
	\end{mdframed}
	
\section{Preliminaries}
\label{sec: whole preliminaries}

	In this section, we first briefly review the basis of flow-based generative models~\citep{lipman2022flow} and then formally define the task of set-structured data generation.

\subsection{Euclidean Flow Matching}
\label{sec: vanilla flow}

	Let the data dimension be $D_{\mathrm{F}}$, the essence of flow matching~\citep{lipman2022flow} is to interpolate between a noise distribution $p_0: \mathbb{R}^{D_{\mathrm{F}}} \rightarrow \mathbb{R}^+$ and the potential data distribution $p_1$, leading to a trajectory of marginal distributions $\{p_{t}\}_{t \in [0, 1]}$. This type of interpolation is realized by a temporal-dependent invertible smooth map (i.e., flow) $\bm{\phi}_t: \mathbb{R}^{D_{\mathrm{F}}} \rightarrow \mathbb{R}^{D_{\mathrm{F}}}, t \in [0, 1]$, satisfying the below ordinary differential equation (ODE):
	\begin{equation}
		\label{eq: vanilla flow ODE}
		\frac{d \bm{\phi}_t(\mathbf{z})}{dt} = \mathbf{v}_{\theta, t}(\bm{\phi}_t(\mathbf{z})), \bm{\phi}_0 = \mathrm{Id},
	\end{equation}
	where vector field $\mathbf{v}_{\theta, t}$ is parameterized by a neural network model and $\mathrm{Id}$ denotes the identity function. The goal is to find a vector field $\mathbf{v}_{\theta, t}$ such that the pushforward of flow $\bm{\phi}_t$ derive the distribution $p_t$ as $\# \bm{\phi}_t [p_0] = p_t$.
	
	The model $\mathbf{v}_{\theta, t}$ is trained by conditional flow matching. Specifically, a conditional flow is predefined as
	\begin{equation}
		\label{eq: vanilla conditional flow}
		\mathbf{z}_{\mathrm{mid}} = \bm{\phi}_t (\mathbf{z} \mid \mathbf{z}_{\mathrm{cond}} ) = (1 - (1 - \zeta) t) \mathbf{z} + t \mathbf{z}_{\mathrm{cond}},
	\end{equation}
	where $\zeta$ is a small constant, with an analytical form of the conditional vector field as
	\begin{equation}
		\label{eq: vanilla conditional vec}
		\mathbf{v}_t(\mathbf{z} \mid \mathbf{z}_{\mathrm{cond}} ) = \frac{\mathbf{z}_{\mathrm{cond}} - (1 - \zeta) \mathbf{z}}{1 - (1 - \zeta) t}.
	\end{equation}
	The loss function is the mean square error (MSE) between this vector field and the model $\mathbf{v}_{\theta, t}$:
	\begin{equation}
		\label{eq: vanilla loss}
		 \mathcal{J}_{\mathrm{E}}  = \mathbb{E}_{t, \mathbf{z}, \mathbf{z}_{\mathrm{cond}}}[\| \mathbf{v}_{\theta, t}(\mathbf{z}_{\mathrm{mid}}) - 	\mathbf{v}_t(\mathbf{z}_{\mathrm{mid}} \mid \mathbf{z}_{\mathrm{cond}} )  \|_2^2],
	\end{equation}
	where $t \sim \mathcal{U}\{0, 1\}$, $\mathbf{z} \sim p_0$, and $\mathbf{z}_{\mathrm{cond}} \sim p_1$. Commonly speaking, the initial distribution $p_0$ is pre-determined as a standard Gaussian $\mathcal{G}(\mathbf{0}, \mathbf{I})$.

\subsection{Problem Formulation}

	An instance of set-structured data $\mathbf{X}$ can be defined as an unordered set of vectors in the Euclidean space: $\{\mathbf{x}_i \in \mathbb{R}^{D_{\mathrm{X}}} \mid i \in [1, N] \cap \mathbb{Z}^+ \}$, where $D_{\mathrm{X}}$ denotes the fixed dimension and $N$ is an uncertain integer. Conventionally, the generation of set-structured data is modeled by the point process~\citep{cox1980point}. In this work, we propose to represent the unordered data $\mathbf{X}$ as some function and aim to model the probability measure of such functions. Specifically, we can uniquely convert any set-structured data $\mathbf{X}$ into a sum of Dirac delta functions:
	\begin{equation}
		\label{eq: delta repr}
		f_{\mathbf{X}} = \sum_{1 \le i \le N} \frac{1}{N} \delta_{\mathbf{x}_i},
	\end{equation}
	in which the term $\delta_{\star}$ is infinite at point $\star$ and vanishes elsewhere. All possible functions of this type form a function space $\mathcal{F}_{\mathrm{delta}}$. To define the probability measure $\mu_{\mathcal{F}_{\mathrm{delta}}}$ on this abstract space, we first suppose that a potential probability space $(\Omega, \mathcal{B}, \mu_{\Omega})$ models the randomness of set $\mathbf{X}$ and random variable $F: \Omega \rightarrow \mathcal{F}_{\mathrm{delta}}$ maps a latent sample $\omega \in \Omega$ to the function representation $f_{\mathbf{X}}$ of some set $\mathbf{X}$. Under this scheme, the measure $\mu_{\mathcal{F}_{\mathrm{delta}}}$ can be derived by the pushforward operation: $\# F [\mu_{\Omega} ]$. 

	In the rest of this paper, we will develop a flow-based generative model that captures the probability measure $\mu_{\mathcal{F}_{\mathrm{delta}}}$. The main difficulties come from the singularities of delta function $\delta_{\star}$ and non-Euclidean space $\mathcal{F}_{\mathrm{delta}}$.

\section{Method: Unordered Flow}
\label{sec: whole method}

	In this part, we introduce a flow-based generative model: \textit{unordered flow}, for set-structured data generation. We will first improve the functional representation $f_{\mathbf{X}}$ of unordered data $\mathbf{X}$, and then extend the flow matching model to the function space $\mathcal{F}_{\mathrm{delta}}$; Finally, we identify the potential unordered data $\widehat{\mathbf{X}}$ from some function $\widehat{f}_{\star}$.

\subsection{Relaxation of the Delta Representation}

	While the delta representation $f_{\mathbf{X}}$ of set-structured data $\mathbf{X}$ has several appealing properties (e.g., simplicity and uniqueness), it is in fact not easy to handle delta function $\delta_{\star}$ both in practice (e.g., unboundedness) and mathematical analysis (e.g., discontinuity). 
	
	\paragraph{Mixture representation with adaptive variances.} To address the singularities of delta function $\delta_{\star}$, we relax it with the Gaussian approximation: $\mathcal{G}(\star, \epsilon^2 \mathbf{I})$, leading to a continuous representation of unordered data $\mathbf{X}$ as
	\begin{equation}
			f_{\mathbf{X}, \epsilon} = \sum_{1 \le i \le N} \frac{1}{N} \mathcal{G}(\mathbf{x}_i, \epsilon^2 \mathbf{I}),
	\end{equation}
	where $\epsilon \in \mathbb{R}^+$ is some small constant. A concern regarding this new representation is that: no matter how small the constant $\epsilon$ is, there might still exist two points $\mathbf{x}_i, \mathbf{y}_j, i \neq j$ whose distance $\|\mathbf{x}_i - \mathbf{y}_j \|_2$ is of the same scale as $\epsilon$. In that case, the Gaussian $\mathcal{G}(\mathbf{x}_i, \epsilon^2 \mathbf{I})$ will fail to locally approximate delta function $\delta_{\mathbf{x}_i}$, due to the effect of another closely located Gaussian $\mathcal{G}(\mathbf{x}_j, \epsilon^2 \mathbf{I})$. A simple solution to this issue is to adaptively rescale the Gaussian variance in terms of the point-to-point distance:
	\begin{equation}
		\label{eq: sigma repr}
		\left\{\begin{aligned}
			f_{\mathbf{X}, \sigma(\epsilon)} & = \sum_{1 \le i \le N} \frac{1}{N} \mathcal{G}(\mathbf{x}_i, \sigma_i(\epsilon)^2 \mathbf{I}) \\
			\sigma_i(\epsilon) & = \epsilon \ln \big(1 + \min_{j \neq i} \| \mathbf{x}_i - \mathbf{x}_j \|_2 \big)
		\end{aligned}\right..
	\end{equation}
	In this way, a point $\mathbf{x}_i$ that is close to another will correspond to a small Gaussian variance $\sigma_i$, so its approximation $\mathcal{G}(\cdot)$ has a limited impact on other Gaussians.
	
	\paragraph{Nice properties.} In the following, we will first see that the mixture representation $f_{\mathbf{X}, \sigma}(\epsilon)$ of unordered data $\mathbf{X}$ is proper for use if we set a small constant $\epsilon$.
	
	\begin{proposition}[Convergent Representation] Suppose that the set of vectors $\mathbf{X}$ is finite, then the limit:
		\begin{equation}
			\lim_{\epsilon \rightarrow 0} f_{\mathbf{X}, \sigma(\epsilon)} = f_{\mathbf{X}},
		\end{equation}
		will hold in a weak sense, with a linear convergence speed in terms of the Wasserstein distance:
		\begin{equation}
			\mathcal{W}_2(f_{\mathbf{X}, \sigma(\epsilon)}, f_{\mathbf{X}}) = \mathcal{O}( \epsilon \sqrt{D_{\mathrm{P}}} \ln \rho),
		\end{equation}
		where $\rho$ is the diameter of set $\mathbf{X}$.
		
		Notably, if the set $\mathbf{X}$ is generated by some inhomogeneous Poisson process, with a regular intensity function, then it is finite with probability $1$.
		\begin{proof}
			The proof is provided in Appendix~\ref{appendix: proof of valid mixture}.
		\end{proof}
	\end{proposition}

	This conclusion not only shows that the limit of representation $f_{\mathbf{X}, \sigma(\epsilon)}$ is equal to the delta representtion $f_{\mathbf{X}}$, but also verifies that the unordered data $\mathbf{X}$ in our formulation is well-defined: $N < \infty$. Besides, the assumption of inhomogeneous Poisson process can generalize many types of unordered data. For the same reason, \citet{kidger2020neural} adopted it to parameterize neural ODE.

	We define the resulting function space that consists of all possible mixture representation $f_{\mathbf{X}, \sigma}(\epsilon)$ as $\mathcal{F}_{\mathrm{mix}}$, with the probability measure denoted as $\mu_{\mathcal{F}_{\mathrm{mix}}}$.

	\begin{proposition}[Regular Function Space]
		\label{prop: regular space}
		The space $\mathcal{F}_{\mathrm{mix}}$ of mixture representation $f_{\mathbf{X}, \sigma(\epsilon)}$ is contained in the space of square-integrable functions: $\mathcal{L}^2(\mathbb{R}^{D_{\mathrm{X}}}) \supseteq \mathcal{F}_{\mathrm{mix}}$. 
		
		As a result, the probability measure $\mu_{\mathcal{F}_{\mathrm{mix}}}$ is supported on the L2 space. However, this is not the case of the probability measure $\mu_{\mathcal{F}_{\mathrm{delta}}}$ of delta representation $f_{\mathbf{X}}$.
		\begin{proof}
			The proof is provided in Appendix~\ref{appendix: proof of the regular space}.
		\end{proof}	
	\end{proposition}

	The significance of this claim is that the new probability measure $\mu_{\mathcal{F}_{\mathrm{mix}}}$ resides in a nice function space: L2. For example, this space can accommodate the Gaussian measure (i.e., infinite-dimensional Gaussian) and has a naturally defined inner product:
	\begin{equation}
		\label{eq: square inner product}
		\langle g, h \rangle_{\mathrm{L2}} = \int g(\mathbf{y}) h(\mathbf{y}) d\mathbf{y},
	\end{equation}
	which facilitates mathematical analysis.
	
	\begin{mdframed}[leftmargin=0pt, rightmargin=0pt, innerleftmargin=10pt, innerrightmargin=10pt, skipbelow=0pt]
		\textbf{\emph{Takeaway}}: The mixture function $f_{\mathbf{X}, \sigma(\epsilon)}$ can represent unordered data $\mathbf{X}$ well, with its probability measure $\mu_{\mathcal{F}_{\mathrm{mix}}}$ supported on the nice L2 space $\mathcal{L}^2(\mathbb{R}^{D_{\mathrm{X}}})$, paving the way for generative modeling.
	\end{mdframed}

\subsection{Function-valued Generative Flow}

	Developing a generative model for the mixture representation $f_{\mathbf{X}, \sigma(\epsilon)}$ seems not trivial, as it is a continuous function, instead of a finite-dimensional vector. However, despite some minor differences in notation, both flow-based and diffusion-based generative models can be easily extended to the infinite-dimensional setting. The foundation of such extensions was rigorously built by many previous works~\citep{kuo2006gaussian,williams2006gaussian,lim2023score,pidstrigach2023infinite,pmlr-v238-kerrigan24a}.

	\paragraph{Generalized flow on the L2 function space.} The L2 space $\mathcal{L}^2(\mathbb{R}^{D_{\mathrm{X}}})$, which supports the probability measure $\mu_{\mathcal{F}_{\mathrm{mix}}}$ (as indicated in Proposition~\ref{prop: regular space}), can be paired with inner product $\langle \cdot, \cdot \rangle_{\mathrm{L2}}$ (i.e., Eq.~(\ref{eq: square inner product})) to form a Hilbert space $\mathcal{H}_{\mathrm{L2}}$. Similar to the Euclidean case (i.e., Sec.~\ref{sec: vanilla flow}), the core of flow matching in space $\mathcal{H}_{\mathrm{L2}}$ is a function-valued vector field $\mathbf{u}_{\theta, t}: \mathcal{H}_{\mathrm{L2}} \rightarrow \mathcal{H}_{\mathrm{L2}}, t \in [0, 1]$ parameterized by neural networks (e.g., Fourier neural operator~\citep{li2021fourier}). Through an ``ODE" that has a very similar form to the Euclidean case (i.e., Eq.~(\ref{eq: vanilla flow ODE})), this vector field $\mathbf{u}_{\theta, t}$ can be used to generate the flow as
	\begin{equation}
		\label{eq: hilbert vec field}
		\frac{d}{dt}[\bm{\varphi}_t(h)] = \mathbf{u}_{\theta, t}(\bm{\varphi}_t(h) ), \bm{\varphi}_0 = \mathrm{Id},
	\end{equation}
	where $d/dt[\cdot], \mathrm{Id}$ are respectively the generalized differential~\citep{agarwal1998difference}) and identity operators on the Hilbert space. As expected, the pushforward $\#$ of this flow $\bm{\varphi}_t$ interpolates between a Gaussian measure $\eta_0 = \mathcal{G}(\mathbf{0}, \bm{\Gamma})$ at the time step $t = 0$ and the target measure $\eta_1 = \mu_{\mathcal{F}_{\mathrm{mix}}}$ at step $t = 1$ as $\eta_t = \# \bm{\varphi}_t [\eta_0]$. A difference is that the mean and covariance coefficients of Gaussian $\mathcal{G}$ is a zero function and a well-defined (e.g., symmetric) linear operator $\bm{\Gamma}: \mathcal{H}_{\mathrm{L2}} \rightarrow \mathcal{H}_{\mathrm{L2}}$.
	
	\paragraph{Training with conditional flow.} Following the Euclidean case (i.e., Eq.~(\ref{eq: vanilla conditional flow})), the training of model $\mathbf{u}_{\theta, t}$ also relies on a conditional flow, which is predefined as
	\begin{equation}
		\label{eq: hilbert cond flow}
		h_{\mathrm{mid}} = \bm{\varphi}_t(h \mid h_{\mathrm{cond}}) = (1 - (1 - \zeta) t) h + t h_{\mathrm{cond}},
	\end{equation}
	with a closed-form conditional vector field:
	\begin{equation}
		\label{eq: hilbert cond vec}
		\mathbf{u}_t(h \mid h_{\mathrm{cond}}) = \frac{h_{\mathrm{cond}} - (1 - \zeta) h}{1 - (1 - \zeta) t}.
	\end{equation}
	Lastly, analogous to Euclidean case $\mathcal{J}_{\mathrm{E}}$ (i.e., Eq.~(\ref{eq: vanilla loss})), a MSE-like loss can be derived as
	\begin{equation}
		\label{eq: hilbert loss}
		\mathcal{J}_{\mathrm{H}} = \mathbb{E} [ \| \mathbf{u}_{\theta, t}(h_{\mathrm{mid}}) - \mathbf{u}_t(h_{\mathrm{mid}} \mid h_{\mathrm{cond}})  \|_{\mathrm{L2}}^2 ],
	\end{equation}
	where the expectation $\mathbb{E}$ is taken over $t \sim \mathcal{U}\{0, 1\}, h \in \eta_0, h_{\mathrm{cond}} \in \eta_1$ and the squared norm $\|\star\|_{\mathrm{L2}}^2$ is induced the inner product as $\langle \star, \star \rangle_{\mathrm{L2}}$.
	
	\begin{mdframed}[leftmargin=0pt, rightmargin=0pt, innerleftmargin=10pt, innerrightmargin=10pt, skipbelow=0pt]
		\textbf{\emph{Note}}:  The arrangement of this subsection is almost the same as Sec.~\ref{sec: vanilla flow} for better understanding. For example, the above Eq.~(\ref{eq: hilbert vec field}), Eq.~(\ref{eq: hilbert cond flow}), Eq.~(\ref{eq: hilbert cond vec}), Eq.~(\ref{eq: hilbert loss}) respectively correspond to the previous Eq.~(\ref{eq: vanilla flow ODE}), Eq.~(\ref{eq: vanilla conditional flow}), Eq.~(\ref{eq: vanilla conditional vec}), Eq.~(\ref{eq: vanilla loss}) in the preliminary.
	\end{mdframed}
	
\subsection{Gradient-based Inverse Transform}
\label{sec: inverse transform}

	A key concern is that, while we can generate mixture representation $\widehat{f}_{\star, \sigma(\epsilon)}$ from a trained \textit{unordered flow} model $\mathbf{u}_{\theta, t}$, it is not obvious how to extract the potential unordered data $\widehat{\mathbf{X}} \subset \mathbb{R}^{D_{\mathbf{X}}}$ hiding in the input domain of this continuous function: $\widehat{f}_{\star, \sigma(\epsilon)}: \mathbb{R}^{D_{\mathrm{X}}} \rightarrow \mathbb{R}$.
	
	\paragraph{Gradient-based search.} If the constant $\epsilon$ is sufficiently small, then the unordered data $\widehat{\mathbf{X}}$ can be approximately regarded as the set of local maxima of function $\widehat{f}_{\star, \sigma(\epsilon)}$. Therefore,  a naive idea is to first start from a random point  $\mathbf{y} \in \mathbb{R}^{D_{\mathrm{X}}}$ and then perform gradient ascent to iteratively update it until converging to some point in set $\widehat{\mathbf{X}}$. Specifically, suppose that the set $\widehat{\mathbf{X}}$ is contained in a bounded region $\mathcal{X} \subset \mathbb{R}^{D_{\mathrm{X}}}$, we sample a number of initial particles $\mathbf{Y}^{(1)} = \{\mathbf{y}_i^{(1)}\}_{1 \le i \le M}, M \in \mathbb{N}^+$ from the region $\mathcal{X}$ and iterate particle updates as
	\begin{equation}
		\label{eq: gradient-based search}
		\left\{\begin{aligned}
			\mathbf{y}_i^{(s + 1)} & = \mathbf{y}_i^{(s)} + \alpha \nabla \widehat{f}_{\star, \sigma(\epsilon)}(\mathbf{y}_i^{(s)} )  \\
			\mathbf{Y}^{(s + 1)} & = \{ \mathbf{y}_i^{(s + 1)} \mid \mathbf{y}_i^{(s)} \in \mathbf{Y}^{(s)} \}
		\end{aligned}\right.,
	\end{equation}
	where superscript $s$ is counted $S_{\mathrm{grad}} \in \mathbb{N}^+$ times and step size $\alpha$ is a small number. Empirically, we find that this gradient-based method is very effective, except that it might not work well when the constant $\epsilon$ is improperly large. In that case, there might exist some \textit{noisy peaks}, which are still a local maximum of function $f_{\mathbf{X}, \sigma(\epsilon)}$ but do not belong to the point set $\mathbf{X}$, misguiding the gradient-based search process. The same problem might also occur when the model $\mathbf{u}_{\theta, t}$ is not well trained, generating a representation $\widehat{f}_{\star, \sigma(\epsilon)}$ that are not mixture-like. For example, the function frequently fluctuates in a certain region, resulting in many ``misleading" local maximum.
	
	\paragraph{Langevin warm-up.} A useful perspective to handle \textit{noisy peaks} is that the mixture representation $f_{\mathbf{X}, \sigma(\epsilon)}$ actually forms a certain density function:
	\begin{equation}\nonumber
		\begin{aligned}
			& \int f_{\mathbf{X}, \sigma(\epsilon)}(\mathbf{y}) d\mathbf{y} = \int \Big( \sum_{1 \le i \le N} \frac{1}{N} \mathcal{G}(\mathbf{y}; \mathbf{x}_i, \sigma_i(\epsilon)^2 \mathbf{I}) \Big) d\mathbf{y} \\
			& = \frac{1}{N} \sum_{1 \le i \le N} \Big( \int \mathcal{G}(\mathbf{y}; \cdot) d\mathbf{y} \Big) = \frac{1}{N}  \sum_{1 \le i \le N} 1 = 1,
		\end{aligned}
	\end{equation}
	and those undesired peaks are with much lower densities than the points in set $\mathbf{X}$. Therefore, we propose to apply Langevin dynamics~\citep{coffey2012langevin} to warm up the uniformly initialized points $\mathbf{Y}^{(1)}$ before performing gradient-based search (i.e., Eq.~(\ref{eq: gradient-based search})):
	\begin{equation}
		\label{eq: langevin iteration}
		\mathbf{y}_i^{(s + 1)} = \mathbf{y}_i^{(s)} + \beta \nabla \ln \widehat{f}_{\star, \sigma(\epsilon)}(\mathbf{y}_i^{(s)} ) + \sqrt{2\beta} \mathbf{z}_i^{(s)},
	\end{equation}
	where $i \in [1, M]$ and $\mathbf{z}_i^{(s)} \sim \mathcal{G}(\mathbf{1}, \mathbf{I})$ is a standard Gaussian noise. The use of this warm-up is to attract random points $\mathbf{Y}^{(1)}$ to high density areas. We have explained why Langevin dynamics can achieve this goal in Appendix~\ref{appendix: lgvin dynamics}. Simply put, Langevin dynamics is a type of Markov chain Monte Carlo (MCMC)~\citep{andrieu2003introduction}, which can converge into a desired distribution.
	
	During \textit{Langevin warm-up}, the superscript $s$ counts from $0$ to a given number $S_{\mathrm{lgvin}} \in \mathbb{N}^+$, and after that, it continues to increase to $S_{\mathrm{lgvin}} + S_{\mathrm{grad}}$ in the gradient-based search process. We denote the final particle set $\mathbf{Y}^{(S_{\mathrm{lgvin}} + S_{\mathrm{grad}})}$ as $\widehat{\mathbf{Y}} =  \{\widehat{\mathbf{y}}_i\}_{1 \le i \le M}$ for convenience.
	
	\paragraph{De-duplication as noise filtering.} The number of particles $M$ should be very large so that the search result $\widehat{\mathbf{Y}}$ can fully cover the potential point set $\widehat{\mathbf{X}}$, though this will cause redundancy: different particles converge to the same point. On the other hand, considering that the search result $\widehat{\mathbf{Y}}$ might contain noises in practice, such redundancy is useful because it indicates how unlikely a particle is noisy, based on whether it is near many other particles.
	
	In this spirit, a simple but very effective de-duplication procedure is designed as follows:
	\begin{enumerate}
		\item  Firstly, we merge close particles in the search result $\widehat{\mathbf{Y}}$ through single-pass clustering~\citep{papka1998line,behnezhad2023single}, resulting in a collection of disjoint particle groups;
		\item Then, we filter out the groups that are small in size, since they are not reliable;
		\item Lastly, we take an average of each group, and all such numbers form the set $\widehat{\mathbf{X}}$.
	\end{enumerate}
	The clustering method in the first step is also very simple. Specifically, we pick up one particle at a time, adding it to the closest group, or initiate a new group that includes the particle if existing groups are all distant.
	
	\begin{algorithm}[t]
		\caption{Inverse Transform: $\widehat{f}_{\star, \sigma(\epsilon)} \mapsto \widehat{\mathbf{X}}$}
		\label{alg: inverse transform}
		\KwIn{Function $\widehat{f}_{\star, \sigma(\epsilon)}$ generated by model $\mathbf{u}_{\theta, t}$}
		\KwOut{Unorded set of vectors: $\widehat{\mathbf{X}}$}
		Sample a particle set $\mathbf{Y}^{(1)}$ from region $\mathcal{X}$ \\
		\For{$s = 1, 2, \cdots S_{\mathrm{lgvin}} - 1$}{
			Warm-up step: $\mathbf{Y}^{(s)} \mapsto \mathbf{Y}^{(s + 1)}$, with Eq.~(\ref{eq: langevin iteration})
		}
		\For{$s = S_{\mathrm{lgvin}}, S_{\mathrm{lgvin}}+ 1, \cdots, S_{\mathrm{lgvin}} + S_{\mathrm{grad}} -1$}{
			Search step: $\mathbf{Y}^{(s)} \mapsto \mathbf{Y}^{(s + 1)}$, with Eq.~(\ref{eq: gradient-based search})
		}
		Merge particles: $\widehat{\mathbf{Y}} = \mathbf{Y}^{(S_{\mathrm{lgvin}} + S_{\mathrm{grad}})} \mapsto \widehat{\mathbf{X}}$
	\end{algorithm}
	
\section{Related Work}
\label{sec: related work}

	There are a substantial number of papers in the literature centering around the topic of unordered data generation. We classify these papers into three categories, and respectively discuss them as follows.

	\paragraph{Classical methods based on point processes.} Conventionally in probability theory, set-structured (i.e., unordered) data are modeled by the point process~\citep{cox1980point}. The core of that framework is the intensity function, indicating how likely an event will occur. For example, the famous Poisson process~\citep{kingman1992poisson} corresponds to a constant intensity function, while its inhomogeneous version means that the function can be arbitrarily valued. More complex models include the Cox process~\citep{cox1955some} that sets the intensity function to be stochastic and the Hawkes process~\citep{hawkes1971spectra} that conditions the intensity function on observed events to capture cross-event dependencies (e.g., self-excitation). The intensity function can be parameterized by neural networks~\citep{omi2019fully} or kernel mixture~\citep{okawa2019deep}. For example, \citet{mei2017neural} applied a variant of LSTM~\citep{graves2012long} to parameterize the Hawkes processes, while \citet{zhang2020self,zuo2020transformer} chose Transformer~\citep{vaswani2017attention}. In the special case of temporal point processes, the data can in fact be ordered in terms of the timestamp. Therefore, some previous works~\citep{du2016recurrent,Shchur2020Intensity-Free} directly learned the time interval distribution, without modeling the intensity function.
	
	\begin{figure*}
		\centering
		\begin{subfigure}[b]{0.245\textwidth}
			\includegraphics[width=\textwidth]{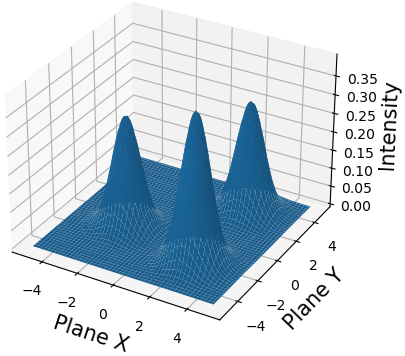}
			\caption{The Poisson intensity that generates the training data.}
			\label{subfig: inhomo-1}
		\end{subfigure}
		\hfill
		\begin{subfigure}[b]{0.245\textwidth}
			\includegraphics[width=\textwidth]{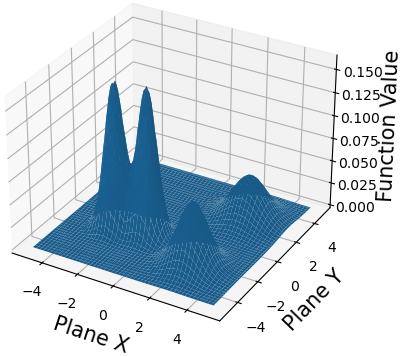}
			\caption{A sample of mixture representation from \textit{unordered flow}.}
			\label{subfig: inhomo-2}
		\end{subfigure}
		\hfill
		\begin{subfigure}[b]{0.245\textwidth}
			\includegraphics[width=\textwidth]{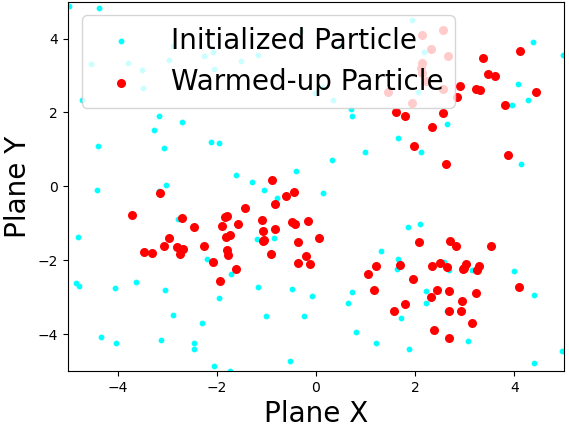}
			\caption{Initial particles and their distribution after warm-up.}
			\label{subfig: inhomo-3}
		\end{subfigure}
		\hfill
		\begin{subfigure}[b]{0.245\textwidth}
			\includegraphics[width=\textwidth]{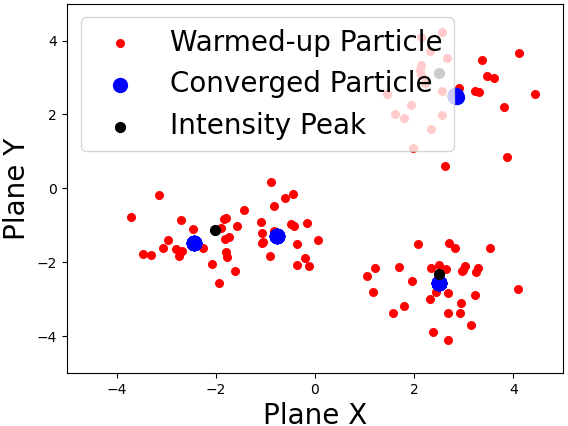}
			\caption{Converged particles after the gradient-based search.}
			\label{subfig: inhomo-4}
		\end{subfigure}
		\caption{Our \textit{unordered flow} model $\mathbf{u}_{\theta, t}$ applied to an inhomogeneous Poisson process. The left two subfigures illustrates  the Poisson intensity function, with a mixture representation $\widehat{f}_{\star, \sigma(\epsilon)}$ generated from the model. The right two respectively depict the warm-up of uniformly initialized particles $\mathbf{Y}^{(1)}$ and their convergence to the point set $\widehat{\mathbf{X}}$.}
		\label{fig: inhomo plots}
	\end{figure*}
	
	For this class of methods, they are either of limited expressiveness (e.g., Hawkes Process) or suppose that the set-like data still have some temporal dimension (i.e., not truly unordered)~\citep{zuo2020transformer}, which is less general than our proposed \textit{unordered flow}.
	
	\paragraph{Methods based on deep generative models.} As deep generative models (e.g., GAN~\citep{goodfellow2020generative}) have gained great popularity, there is a surge of interest adapting those techniques to set-structured data generation~\citep{yang2019pointflow,yang2022conditional}. For example, \citet{bilovs2021scalable} learned a permutation invariant density distribution through normalizing flow~\citep{kingma2018glow}, and \citet{xie2021generative} designed a type of energy-based model~\citep{lecun2006tutorial} for point cloud data, which are inherently unordered. With the emergence of diffusion-based generative model, there are also some explorations~\citep{luo2021diffusion,lyu2022a} that directly applied diffusion models to unordered data (e.g., spatial coordinates), without considering their permutation invariance.
	
	 Notably, in the case where a temporal dimension exists in the data format, some works~\citep{kidger2020neural,chen2021neural,yuan2023spatio} adopted neural ODE~\citep{chen2018neural} or Transformer to model the irregularly sampled events, with diffusion models or normalizing flow learning the spatial distribution. 

	\paragraph{Very recent diffusion-like baselines.} Until recently, there are  a few attempts introducing generative models for unordered data, which mimic how diffusion models work: adding noise and denoising. For example, \citet{ludke2023add,ludke2024unlocking} reformulated the diffusion process as randomly excluding original points and adding noisy ones. This type of methods in fact differ from the conventional diffusion models (which are based on heat equations~\citep{sohl2015deep}) in principle, involving many discrete operations (e.g., thinning). More closely related to this paper, \citet{chen2023learning} introduced JUMP that generalizes the diffusion process from continuous spaces to integer sets, and \citet{bilovs2023modeling} proposed to regard irregular time series as a continuous function for generative modeling. However, those baselines in fact assume that the data exhibits a certain order (e.g., temporal dimension), which are inapplicable to truly unordered data (e.g., point cloud).

\section{Experiments}
\label{sec: whole experiment}

	In this section, we aim to verify the effectiveness of our \textit{unordered flow} model. We show that it performs well on both synthetic and real datasets, outperforming previous baselines. Our extensive experiments (e.g., ablation studies) also indicate the indispensability of different modules in the model. We will first present the experiment setup in the next subsection, with minor details in Appendix~\ref{appendix: experiment details}.
	
\subsection{Settings}

	\paragraph{Datasets.} We adopt both synthetic and real-world datasets for experiments. Specifically, we synthesize training data from two common point processes:  one is a collection of point sets sampled from a 2-dimensional inhomogeneous Poisson process, with its intensity function as a squared-exponential mixture, while the other is drawn from the Hawkes process, with an exponential kernel to parameterize its intensity function.
	
	For the real-world datasets, we follow previous works  \citep{chen2018neural,ludke2024unlocking} to adopt three real-world datasets of point sets: Japan Earthquakes~\citep{usgs2020earthquake}, New Jersey COVID-19 Cases~\citep{data/nytimes_covid19}, and Citibike Pickups~\citep{citi_bike}. The pre-processing and splits of these datasets are consistent with \citet{chen2021neural}.

	\paragraph{Evaluation metrics.} Following a number of previous works~\citep{shchur2020fast,ludke2023add,ludke2024unlocking}, we adopt two evaluation metrics: 1) \textit{S-WStein}, the Wasserstein distance~\citep{ramdas2017wasserstein} that measures the distributional discrepancy between the size of generated point sets and that of real sets; 2) \textit{D-MMD}, the maximum mean discrepancy measure (MMD)~\citep{gretton2012kernel}  that quantifies the distribution gap between two point processes. Note that we did not include negative log-likelihood (NLL) because many types of generation models (e.g., GAN) are not a density estimator and there are many studies~\citep{Theis2016a,shchur2020fast} pointing out that it is in fact not a proper metric. The lower both metrics are, the better the tested model performs.

	\begin{figure*}
		\centering
		\begin{subfigure}[b]{0.245\textwidth}
			\includegraphics[width=\textwidth]{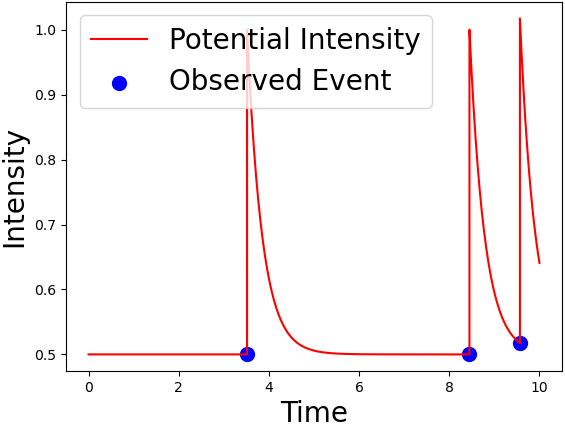}
			\caption{A case of point set for training, with its Hawkes intensity.}
			\label{subfig: hawkes-1}
		\end{subfigure}
		\hfill
		\begin{subfigure}[b]{0.245\textwidth}
			\includegraphics[width=\textwidth]{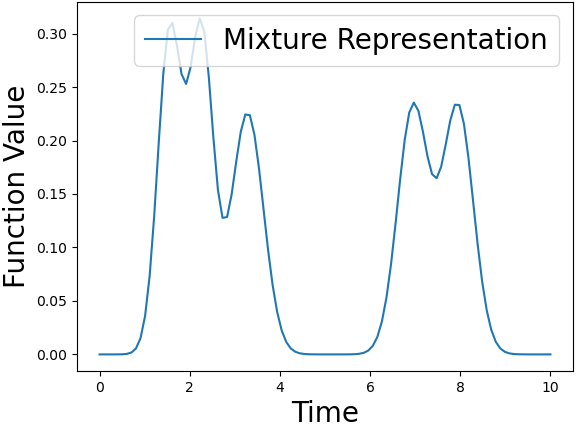}
			\caption{A sample of mixture representation drawn from our model.}
			\label{subfig: hawkes-2}
		\end{subfigure}
		\hfill
		\begin{subfigure}[b]{0.245\textwidth}
			\includegraphics[width=\textwidth]{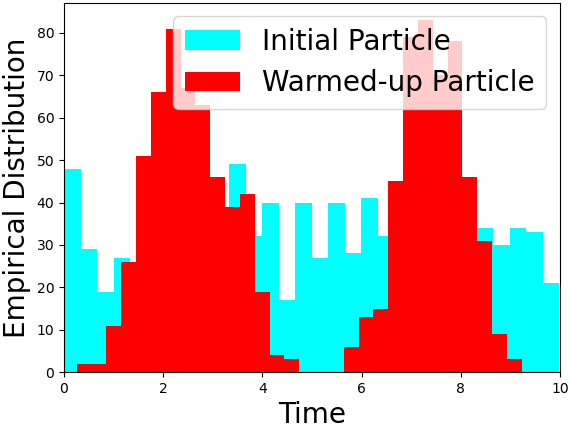}
			\caption{The distributions of initial and warmed-up particles.}
			\label{subfig: hawkes-3}
		\end{subfigure}
		\hfill
		\begin{subfigure}[b]{0.245\textwidth}
			\includegraphics[width=\textwidth]{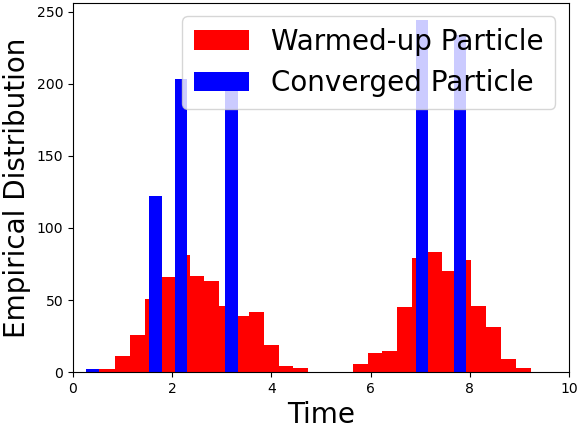}
			\caption{The particle distribution after gradient-based search.}
			\label{subfig: hawkes-4}
		\end{subfigure}
		\caption{Results from our \textit{unordered flow} model $\mathbf{u}_{\theta, t}$ on the Hawkes process. The left two subfigures illustrate a point set $\mathbf{X}$ in the training set, with a mixture representation $\widehat{f}_{\star, \sigma(\epsilon)}$ sampled from the model. The right two show the empirical distributions of particles, including their initial  $\mathbf{Y}^{(1)}$, warmed-up $\mathbf{Y}^{(S_{\mathrm{lgvin}})}$, and converged versions $\widehat{\mathbf{Y}}$.}
		\label{fig: hawkes plots}
	\end{figure*}
	
	\begin{table*}[t]
		\centering
		\begin{tabular}{c|cc|cc|cc}
			
			\hline
			\multirow{2}{*}{Method} & \multicolumn{2}{c|}{Earthquakes}  & \multicolumn{2}{c}{COVID-19} & \multicolumn{2}{|c}{Citibike} \\
			\cline{2-7}
			& S-WStein & D-MMD & S-WStein & D-MMD   &  S-WStein  & D-MMD  \\
			
			\hline
			Log-Gaussian Cox Process & $0.047$ &  $0.214$ & $0.209$ & $0.340$ & $0.104$ & $0.336$  \\
			
			Permutation-invariant Normalizing Flow & $0.043$ & $0.191$ & $0.185$ & $0.271$ &$0.071$ & $0.105$  \\
			
			Energy-based Generative PointNet & $0.040$ & $0.185$ & $0.201$ & $0.253$ & $0.063$ & $0.091$ \\
			
			Point Set Diffusion& $0.038$ & $0.173$ & $0.199$ & $0.268$ & $0.056$ & $0.092$ \\
			
			\hline
			Our Mode: \textit{Unordered Flow} & $\mathbf{0.023}$ & $\mathbf{0.125}$ & $\mathbf{0.153}$ & $\mathbf{0.213}$  & $\mathbf{0.041}$ & $\mathbf{0.079}$ \\
			\hline
			
		\end{tabular}
		\caption{The performances of our model and baselines on three real-world datasets. The two metrics: \textit{S-WStein} and \textit{D-MMD}, respectively measure the distributional discrepancies between the generated and real point sets in terms of the set size and point position. Plus, the results from our implemented models are averaged over 10 runs.}
		\label{tab: main experiment}
	\end{table*}

	\paragraph{Baselines.} We adopt multiple key baselines for comparison. Specifically, they are log-Gaussian Cox process~\citep{moller1998log}: a classical point process model, Generative PointNet~\citep{xie2021generative} and order-insensitive Normalizing Flow~\citep{bilovs2021scalable}: deep generative models extended to unordered data, and Point Set Diffusion~\citep{ludke2024unlocking}: a diffusion-like generative models but with a different working principle from typical diffusion models. Each baseline adopted here has been comprehensively discussed in Sec.~\ref{sec: related work}. While the results of Cox process and Point Set Diffusion are copied from \citet{ludke2024unlocking}, those of other two baselines come from our implementation.

\subsection{Proof-of-concept Studies}

	We aim to first verify whether our  \textit{unordered flow} model can approximate two common point processes: inhomogeneous Poisson process and Hawkes process, with an emphasize to show how its workflow runs in practice.
	
	The results on the Poisson process are shown in Fig.~\ref{fig: inhomo plots}. From Subfig.~\ref{subfig: inhomo-2}, we can see that our model generates a function $\widehat{f}_{\star, \sigma(\epsilon)}$ that is consistent with the form of training data (i.e., mixture representation $f_{\mathbf{X}, \epsilon}$), with density peaks located in the high-intensity areas shown in Subfig.~\ref{subfig: inhomo-1}. As introduced in Sec.~\ref{sec: inverse transform}, to decode the potential point set $\widehat{\mathbf{X}}$ from the sampled function $\widehat{f}_{\star, \sigma(\epsilon)}$, we first uniformly sample a number of particles $\mathbf{Y}^{(1)}$, with \textit{Langevin warm-up} moving them towards high-density regions, which is exactly the case as illustrated in Subfig.~\ref{subfig: inhomo-3}. Finally, Subfig.~\ref{subfig: inhomo-4} shows that the warmed-up particles  $\mathbf{Y}^{(S_{\mathrm{lgvin}})}$ perfectly converge to the density peaks of $f_{\mathbf{X}, \epsilon}$ (i.e., the hidden point set $\widehat{\mathbf{X}}$) after the gradient-based search.
	
	The results on the Hawkes process as shown in Fig.~\ref{fig: hawkes plots} are also successful. Similarly, the generated function from our \textit{unordered flow} is mixture-like (Subfig.~\ref{subfig: hawkes-2}), and the initial particles are warmed-up well (Subfig.~\ref{subfig: hawkes-3}), with convergence to the density peaks of the sampled function (Subfig.~\ref{subfig: hawkes-4}). Both our results on the Poisson and Hawkes processes demonstrate that every module of \textit{unordered flow} is effective, and all these modules form a well-performing generative model for point processes.

\subsection{Evaluation on Real Datasets}

	\begin{table*}
		\centering
		\begin{tabular}{c|cc|cc}
			\hline
			\multirow{2}{*}{Method} & \multicolumn{2}{c|}{Earthquakes}  & \multicolumn{2}{c}{COVID-19} \\
			\cline{2-5}
			& S-WStein& D-MMD & S-WStein & D-MMD \\
			
			\hline
			Our Mode: \textit{Unordered Flow} & $\mathbf{0.023}$ & $\mathbf{0.125}$ & $\mathbf{0.153}$ & $\mathbf{0.213}$ \\ \hline
			w/o Adaptive Variance (i.e., Eq.~(\ref{eq: sigma repr})) & 0.028 & 0.136 & 0.185 & 0.232 \\
			w/o \textit{Langevin Warm-up} (i.e., Eq.~(\ref{eq: langevin iteration})) & 0.031 & 0.146 & 0.176 & 0.239 \\
			w/o Noisy Peak Filtering & 0.033 & 0.143 & 0.181 & 0.229  \\
			\hline
		\end{tabular}
		\caption{The results from our ablation studies. Each of the last three rows shows the changed model performance after a specific module (e.g., \textit{Langevin warm-up}) excluded from \textit{unordered flow}.}
		\label{tab: ablation studies}
	\end{table*}

	\begin{figure*}
		\centering
		\begin{subfigure}[b]{0.28\textwidth}
			\includegraphics[width=\textwidth]{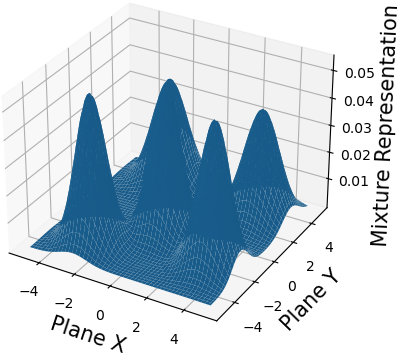}
			\caption{A mixture representation sampled from \textit{unordered flow}.}
			\label{subfig: warmup-1}
		\end{subfigure}
		\hfill
		\begin{subfigure}[b]{0.33\textwidth}
			\includegraphics[width=\textwidth]{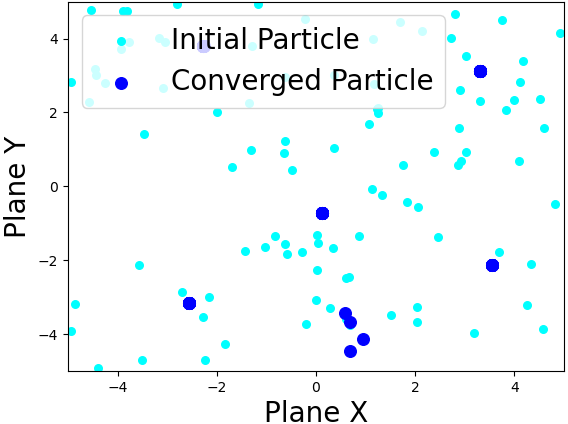}
			\caption{Initial particles converge without warm-up, but some fall into ``noisy peaks".}
			\label{subfig: warmup-2}
		\end{subfigure}
		\hfill
		\begin{subfigure}[b]{0.33\textwidth}
			\includegraphics[width=\textwidth]{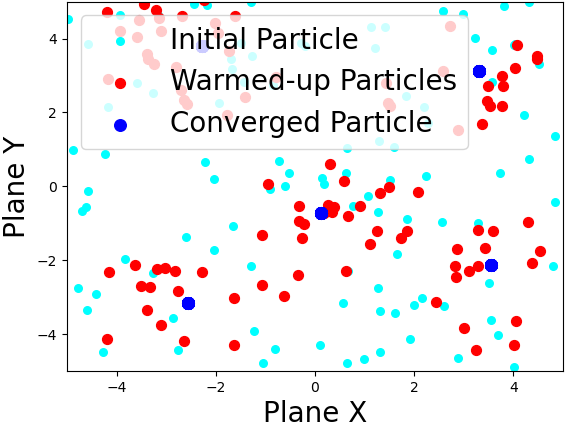}
			\caption{Initial particles are warmed-up well and converge efficiently.}
			\label{subfig: warmup-3}
		\end{subfigure}
		\caption{A case study to show the importance of \textit{Langevin warm-up} in the inverse transform that identifies the point set $\widehat{\mathbf{X}}$ from a mixture representation $\widehat{f}_{\star, \sigma(\epsilon)}$ (i.e., left subfigure). The middle subfigure is the convergence of particles without \textit{Langevin warm-up}, while the right one is with the warm-up.}
		\label{fig: warm-up case}
	\end{figure*}

	We also applied \textit{unordered flow} to real datasets, evaluating its performance in practice. The results are shown in Table~\ref{tab: main experiment}. We can see that our model significantly outperforms previous baselines on all the datasets and in terms of each metric. For example, our model has achieved a much lower \textit{D-MMD} score than the latest baseline (i.e., Point Set Diffusion) by $20.52\%$ on the COVID-19 dataset, which is $39.47\%$ for \textit{S-WStein} on the Earthquakes dataset. These notably improvements over baselines strongly confirm the effectiveness of  \textit{unordered flow}.

	An interesting observation from the table is that a previous energy-based generative baseline (i.e., Generative PointNet) performed competitively with the very recent Point Set Diffusion, which is still built on top of the framework of point processes. For example, their performance gap on the Citibik dataset is only $1.09\%$ in terms of \textit{D-MMD}. This indicates that the direction of applying deep generative models to unordered data is promising.

\subsection{Ablation Studies}

	We have conducted ablation experiments on two real datasets to verify the effectiveness of multiple modules in our \textit{unordered flow} model. The results are shown in Table~\ref{tab: ablation studies}. We can see that every tested module is crucial to our model performance. For example, the last row in the table means not to exclude small groups after clustering converged particles. This leads to a significant increase by $18.30\%$ on the COVID-19 dataset in terms of \textit{S-WStein}.

 \subsection{Case Study}
 
 	While the ablation studies have quantitatively verified that \textit{Langevin warm-up} is essential to the model performance, we showcase this point by a real example.
 	
 	The case is demonstrated in Fig.~\ref{fig: warm-up case}. From Subfig.~\ref{subfig: warmup-2}, we can see that the uniformly initialized particles $\mathbf{Y}^{(1)}$ without \textit{Langevin warm-up} might converge to some ``noisy peaks" (mentioned in Sec.~\ref{sec: inverse transform}) in the bottom center, and this problem can be perfectly addressed by the warm-up trick as shown in Subfig.~\ref{subfig: warmup-3}. Another observation is that \textit{Langevin warm-up} greatly speeds up the particle filtering process. The convergence in Subfig.~\ref{subfig: warmup-3} only takes $\mathcal{S}_{\mathrm{lgvin}} + S_{\mathrm{grad}} = 500$ iterations in total, which is $5000$ in Subfig.~\ref{subfig: warmup-2}.
 
\section{Conclusion}
	
	In this paper, we present \textit{unordered flow}, a type of flow-based generative model for set-structured data generation, filling a gap in the literature that there are no permutation-invariant flow-based or diffusion-based generative models. The core of this technique is to map the unordered data into a mixture representation in the function space, which we prove is accurate and derives a nice probability measure supported on the L2 space, facilitating function-valued flow matching. For the inverse transform that maps a given function back to the unordered data, we propose a particle filtering method, with Langevin dynamics to warm up the particles and gradient ascent to update them until convergence. Extensive experiments on both synthetic and real datasets demonstrate that our \textit{unordered flow} model can generate set-structured data well, significantly outperforming previous key baselines.

%
%


\bibliography{example_paper}
\bibliographystyle{icml2025}

\newpage
\appendix
\onecolumn

\section{Proof: Convergence of the Mixture Representation}
\label{appendix: proof of valid mixture}

	We will first prove the first and last claims, and then go to the second one.

\subsection{The Limit with a Reducible Assumption}

	Since the point set $\mathbf{X}$ is finite (i.e., $N < \infty$), we can infer that there exists a finite number $\rho$ satisfying
	\begin{equation}
		\infty > \rho > \sup_{i \neq j} \| \mathbf{x}_i - \mathbf{x}_j \|_2.
	\end{equation}
	A known fact is that the Gaussian $\mathcal{G}(\mathbf{x}, \epsilon^2 \mathbf{I})$ weakly converges to delta function $\delta_{\mathbf{x}}$ as $\epsilon \rightarrow 0$. Specifically, for any smooth test function $g$, this means: given any $\epsilon_2 > 0$, there always exists a number $\epsilon_1 > 0$ such that
	\begin{equation}
		\Big|\int \mathcal{G}(\mathbf{y}; \mathbf{x}, \epsilon^2 \mathbf{I}) g(\mathbf{y}) d\mathbf{y} - \int \delta_{\mathbf{x}}(\mathbf{y}) g(\mathbf{y}) d\mathbf{y} \Big| < \epsilon_2
	\end{equation}
	holds for any $\epsilon \in (0, \epsilon_1]$. In terms of this fact, given the error $\epsilon_2$, there always exists a finite set of numbers $\{\epsilon_i\}_{i \in [1, N]}$ such that the below inequality:
	\begin{equation}
		\Big|\int (\mathcal{G}(\mathbf{y}; \mathbf{x}_i, \epsilon^2 \rho^2 \mathbf{I}) - \delta_{\mathbf{x}_i}(\mathbf{y})) g(\mathbf{y}) d\mathbf{y} \Big| < \epsilon_2, 
	\end{equation}
	holds for any $i \in [1, N], \epsilon < (0, \epsilon_1]$. Following these inequalities, we have
	\begin{equation}
		\begin{aligned}
			& \Big| \int (f_{\mathbf{X}, \sigma(\epsilon)}(\mathbf{y}) - f_{\mathbf{X}}(\mathbf{y})) g(\mathbf{y}) d\mathbf{y}  \Big| = \Big| \int \Big( \sum_{1 \le i \le N} \frac{1}{N} \mathcal{G}(\mathbf{y}; \mathbf{x}_i, \sigma_i(\epsilon)^2 \mathbf{I}) - \sum_{1 \le i \le N} \frac{1}{N} \delta_{\mathbf{x}_i}(\mathbf{y}) \Big) g(\mathbf{y}) d\mathbf{y}  \Big| \\
			& = \Big| \frac{1}{N} \sum_{1 \le i \le N} \int (\mathcal{G}(\mathbf{y}; \mathbf{x}_i, \sigma_i(\epsilon)^2 \mathbf{I}) - \delta_{\mathbf{x}_i}(\mathbf{y})) g(\mathbf{y}) d\mathbf{y} \Big| \\
			& < \frac{1}{N} \sum_{1 \le i \le N} \Big| \int (\mathcal{G}(\mathbf{y}; \mathbf{x}_i, \sigma_i(\epsilon)^2 \mathbf{I}) - \delta_{\mathbf{x}_i}(\mathbf{y})) g(\mathbf{y}) d\mathbf{y}  \Big| < \frac{1}{N} \sum_{1 \le i \le N} \epsilon_2 = \epsilon_2,
		\end{aligned}
	\end{equation}
	where the last inequality holds because
	\begin{equation}
		\sigma_i(\epsilon) = \epsilon \ln\big( 1 +  \min_{j \neq i} \| \mathbf{x}_i - \mathbf{x}_j \|_2 \big) \le \epsilon \min_{j \neq i} \| \mathbf{x}_i - \mathbf{x}_j \|_2  \le \epsilon \big( \sup_{j \neq k} \| \mathbf{x}_j - \mathbf{x}_k \|_2 \big) = \epsilon \rho.
	\end{equation}
	Therefore, for any smooth test function $g$, we can infer that
	\begin{equation}
		\lim_{\epsilon \rightarrow 0} \int f_{\mathbf{X}, \sigma(\epsilon)}(\mathbf{y}) g(\mathbf{y}) d\mathbf{y}  = \int f_{\mathbf{X}}(\mathbf{y}) g(\mathbf{y}) d\mathbf{y}.
	\end{equation}
	With this result, it is known in the distribution theory~\citep{friedlander1998introduction} that the mixture representation $ f_{\mathbf{X}, \sigma(\epsilon)}(\mathbf{y})$ is equal to the delta representation $f_{\mathbf{X}}(\mathbf{y})$ in a weak sense.
	
	For the second claim, the intensity function $\lambda(\mathbf{y}) \ge 0, \mathbf{y} \in \mathbb{R}^{D_{\mathrm{X}}}$ of a generalized Poisson process is said to be regular if its integral over the (possibly unbounded) support $\mathcal{R} \subseteq  \mathbb{R}^{D_{\mathrm{X}}}$ is finite:
	\begin{equation}
		\Lambda(\mathcal{R}) = \int_{\mathcal{X}} \lambda(\mathbf{y}) d\mathbf{y} < \infty.
	\end{equation}
	A key conclusion from the theory of point processes~\citep{cox1980point} is that the probability of event occurrence $\mathbf{X}$ can be formulated as
	\begin{equation}
		\mathrm{Prob}(\mathbf{X}) = \exp(- \Lambda(\mathcal{R})) \prod_{1 \le i \le N} \lambda(\mathbf{x}_i),
	\end{equation}
	and the number of points $N = |\mathbf{X}|$ is with respect to a Poisson:
	\begin{equation}
		\mathrm{Prob}(N = k) = \exp(-\Lambda(\mathcal{R})) \frac{\Lambda(\mathcal{R})^k}{k!}.
	\end{equation}
	Based on this fact, we can infer that
	\begin{equation}
		\mathrm{Prob}(N < \infty) = \sum_{k \ge 0} \mathrm{Prob}(N = k) =  \exp(-\Lambda(\mathcal{R})) \sum_{k \ge 0} \frac{\Lambda(\mathcal{R})^k}{k!} = 1,
	\end{equation}
	which exactly verifies the second claim in the proposition.
	
\subsection{Analysis of the Convergence Speed}

	The Wasserstein distance~\citep{ramdas2017wasserstein} $\mathcal{W}_2$ between two distributions $\pi_1, \pi_2: \mathbb{R}^{D_{\mathrm{P}}} \rightarrow R$ is defined as
	\begin{equation}\nonumber
		\mathcal{W}_2(\pi_1, \pi_2)^2 = \inf_{\lambda \in \Lambda (\pi_1, \pi_2)} \Big( \int \|\mathbf{x} - \mathbf{y}\|^2_2 d \lambda(\mathbf{x}, \mathbf{y}) \Big),
	\end{equation}
	where $\Lambda$ denotes the set of all possible joint distributions, with marginals the same as its arguments.

	We first show that metric $\mathcal{W}_2$ enjoys certain decomposability. Suppose that there are any two probability measures $\pi_1, \pi_2$ that shape as a mixture. Formally, we formulate them as
	\begin{equation}
		\pi_1 = \sum_{1 \le i \le n} w_{1, i} \pi_{1, i}, \ \ \ \pi_2 = \sum_{1 \le j \le m} w_{2, j} \pi_{2, j}, 
	\end{equation}
	where $\sum_i w_{1, i} = 1$, $\sum w_{2, j} = 1$, and $\pi_{1, i}, \pi_{2, j}$ are also probability measures. We further suppose that the distance $\mathcal{W}_2(\pi_{1, i}, \pi_{2, j} ) < \infty$ is finite for $\forall i, j \in [1, n] \times [1, m]$, then there exists a positive number $c_{i,j}$ that bounds its squared value. By definition we know that there is also a probability measure $\widetilde{\lambda}_{i,j} \in \Gamma(\pi_{1, i}, \pi_{2, j})$ satisfying
	\begin{equation}
		\int \|\mathbf{x} - \mathbf{y}\|^2_2 d \widetilde{\lambda}_{i,j}(\mathbf{x}, \mathbf{y})  < c_{i,j}.
	\end{equation}
	Now, we construct a new probability measure as
	\begin{equation}
		\widetilde{\lambda} = \sum_{1 \le i \le n}  \sum_{1 \le j \le m} w_{1, i} w_{2, j} \widetilde{\lambda}_{i,j}.
	\end{equation}
	It is easy to verify that this new measure $\widetilde{\lambda}$ belong to $\Lambda (\pi_1, \pi_2)$ as
	\begin{equation}
		\begin{aligned}
			& \int \widetilde{\lambda}(\mathbf{x}, \mathbf{y}) d\mathbf{x} = \int \Big(   \sum_{1 \le i \le n}  \sum_{1 \le j \le m} w_{1, i} w_{2, j} \widetilde{\lambda}_{i,j}(\mathbf{x}, \mathbf{y})  \Big) d\mathbf{x} \\
			& =  \sum_{1 \le i \le n}  \sum_{1 \le j \le m} w_{1, i} w_{2, j} \Big( \int \widetilde{\lambda}_{i,j}(\mathbf{x}, \mathbf{y}) d\mathbf{x} \Big) =  \sum_{1 \le i \le n}  \sum_{1 \le j \le m} w_{1, i} w_{2, j}  \pi_{2, j}(\mathbf{y}) = \\
			& \Big( \sum_{1 \le i \le n}  w_{1, i} \Big) \Big( \sum_{1 \le j \le m} w_{2, j} \pi_{2, j}(\mathbf{y}) \Big) =  1 \cdot \pi_{2}(\mathbf{y}) =  \pi_{2}(\mathbf{y}).
		\end{aligned}
	\end{equation}
	Symmetrically, we can also prove that $\int \widetilde{\lambda}(\mathbf{x}, \mathbf{y}) d\mathbf{y} = \pi_1(\mathbf{x})$ using the same derivation strategy. Based on the definition of Wasserstein  metric $\mathcal{W}_2$, we have
	\begin{equation}
		\begin{aligned}
			& \mathcal{W}_2(\pi_1, \pi_2)^2 \le \int \|\mathbf{x} - \mathbf{y}\|^2_2 d \widetilde{\lambda}(\mathbf{x}, \mathbf{y}) = \\
			&  \sum_{1 \le i \le n}  \sum_{1 \le j \le m} w_{1, i} w_{2, j} \int \|\mathbf{x} - \mathbf{y}\|^2_2 d \widetilde{\lambda}_{i,j}(\mathbf{x}, \mathbf{y}) \le  \sum_{1 \le i \le n}  \sum_{1 \le j \le m} w_{1, i} w_{2, j} c_{i,j}.
		\end{aligned}
	\end{equation}
	Now, we let the constant $c_{i,j}$ go to its limit: $c_{i,j} \rightarrow \mathcal{W}_2(\pi_{1, i}, \pi_{2, j} )^2$, leading to
	\begin{equation}
		 \mathcal{W}_2(\pi_1, \pi_2)^2 \le \sum_{1 \le i \le n}  \sum_{1 \le j \le m} w_{1, i} w_{2, j} \mathcal{W}_2(\pi_{1, i}, \pi_{2, j} )^2.
	\end{equation}
	Then, we aim to study the Wasserstein distance $\mathcal{W}_2$ between Gaussian $\pi_1 = \int \mathcal{G}(\mathbf{z}_1, \gamma^2 \mathbf{I})$ and delta distribution $\pi_2 =  \int \delta_{\mathbf{z}_2}$. Here we abuse the symbol $\int$ to denote an integral-like operator: mapping a density function to its measure. Note that the latter is only supported on a single point: $\mathbf{z}_2$. Therefore, there is in fact only one trivial joint measure: $\lambda = \int \mathcal{G}(\mathbf{z}_1, \gamma^2 \mathbf{I}) \cdot \int \delta_{\mathbf{z}_2}$ in space $\Lambda(\cdot)$. In this sense, we have
	\begin{equation}
		\begin{aligned}
			& \mathcal{W}_2(\mathcal{G}(\mathbf{z}_1, \gamma^2 \mathbf{I}), \delta_{\mathbf{z}_2})^2 = \int_{\mathbf{x}} \int_{\mathbf{y}} \|\mathbf{x} - \mathbf{y}\|^2_2 \mathcal{G}(\mathbf{x}; \mathbf{z}_1, \gamma^2 \mathbf{I}) \delta_{\mathbf{z}_2}(\mathbf{y}) d\mathbf{x} d\mathbf{y} \\
			& = \int_{\mathbf{x}}  \mathcal{G}(\mathbf{x}; \mathbf{z}_1, \gamma^2 \mathbf{I})  \Big( \int_{\mathbf{y}} \|\mathbf{x} - \mathbf{y}\|^2_2 \delta_{\mathbf{z}_2}(\mathbf{y}) d\mathbf{y} \Big) d\mathbf{x} = \int_{\mathbf{x}}  \mathcal{G}(\mathbf{x}; \mathbf{z}_1, \gamma^2 \mathbf{I}) \| \mathbf{x} - \mathbf{z}_2 \|^2_2 d\mathbf{x} \\
			& = \mathbb{E}_{\mathbf{x} \sim \mathcal{G}(\mathbf{z}_1, \gamma^2 \mathbf{I})} \Big[  \| \mathbf{x} - \mathbf{z}_2 \|^2_2 \Big] =  \mathbb{E}_{\mathbf{x} \sim \mathcal{G}(\mathbf{0}, \mathbf{I})} \Big[  \|  \gamma \mathbf{x}  + \mathbf{z}_1 - \mathbf{z}_2 \|^2_2 \Big] \\
			& = \gamma^2 \mathbb{E}_{\mathbf{x} \sim \mathcal{G}(\mathbf{0}, \mathbf{I})} \Big[  \|  \mathbf{x}  \|_2^2 \Big] + 2\gamma \mathbb{E}_{\mathbf{x} \sim \mathcal{G}(\mathbf{0}, \mathbf{I})} \Big[(\mathbf{z}_1 - \mathbf{z}_2)^{\top}\mathbf{x} \Big] + \|  \mathbf{z}_1 - \mathbf{z}_2 \|^2_2 =  \gamma^2 D_{\mathrm{P}} + \|  \mathbf{z}_1 - \mathbf{z}_2 \|^2_2 
		\end{aligned}
	\end{equation} 
	Combining the above two key conclusions, we have
	\begin{equation}
		\begin{aligned}
			& \mathcal{W}_2(f_{\mathbf{X}, \sigma(\epsilon)}, f_{\mathbf{X}})^2 \le \sum_{1 \le i, j \le N} \frac{1}{N^2}  \mathcal{W}_2(  \mathcal{G}(\mathbf{x}_i, \sigma_i(\epsilon)^2 \mathbf{I}), \delta_{\mathbf{x}_j}  )^2 \\
			& = \frac{1}{N^2} \sum_{1 \le i, j \le N} \Big(  \sigma_i(\epsilon)^2 D_{\mathrm{P}} + \|  \mathbf{x}_i - \mathbf{x}_j \|^2_2 \Big) \\
			& \le  \frac{1}{N^2}  \sum_{1 \le i, j \le N} \Big ( \epsilon^2 D_{\mathrm{P}}  \min_{k \neq i} \| \mathbf{x}_i - \mathbf{x}_k \|_2^2 + \|  \mathbf{x}_i - \mathbf{x}_j \|^2_2 \Big) \\
			& \le \frac{1}{N^2}  \sum_{1 \le i, j \le N} \rho^2 ( \epsilon^2 D_{\mathrm{P}} + 1) = \rho^2 ( \epsilon^2 D_{\mathrm{P}} + 1).
		\end{aligned}
	\end{equation}
	We can see that the outcome in this derivation has an undesired term $\rho^2$, resulting in a too loose upper bound. To address this problem, we consider a different strategy in this stage: The essence of  Wasserstein distance $\mathcal{W}_2$ is to measure the minimum cost of transforming the probability density from one to another. Therefore, any transform plan that we can propose must incur a cost that is not smaller than the distance.
	
	In this spirit, a transform plan $\widetilde{\pi}$ is to move the whole probability mass (i.e., $1$) of every Gaussian $ \mathcal{G}(\mathbf{x}_i, \sigma_i(\epsilon)^2 \mathbf{I})$ in mixture representation $f_{\mathbf{X}, \sigma(\epsilon)}$ to the Dirac $\delta_{\mathbf{x}_i}$ in delta representation $ f_{\mathbf{X}}$ that has the same center:
	\begin{equation}
		\pi(\mathcal{X}, \mathbf{y} = \mathbf{x}_i) = \frac{1}{N} \int \Big[ \mathcal{G}(\mathbf{x}_i, \sigma_i(\epsilon)^2 \mathbf{I}) \Big](\mathcal{X}), \ \ \  	\pi(\mathcal{X}, \mathbf{y} \notin \mathbf{X}) = 0 
	\end{equation}
	where $\mathcal{X}$ is a Borel set in $\mathbb{R}^{D_{\mathrm{P}}}$. Considering the previous conclusion, we have
	\begin{equation}
		\begin{aligned}
			& \mathcal{W}_2(f_{\mathbf{X}, \sigma(\epsilon)}, f_{\mathbf{X}})^2 \le \int \|\mathbf{x} - \mathbf{y}\|^2_2 d \widetilde{\lambda}(\mathbf{x}, \mathbf{y}) = \int_{\mathbf{x}} \int_{\mathbf{y}} \|\mathbf{x} - \mathbf{y}\|^2_2 \Big(  \frac{1}{N} \sum_{1 \le i \le N} \mathcal{G}(\mathbf{x}; \mathbf{x}_i, \sigma_i(\epsilon)^2 \mathbf{I}) \delta_{\mathbf{x}_i}(\mathbf{y}) \Big) d\mathbf{x} d\mathbf{y} \\
			& = \frac{1}{N} \sum_{1 \le i \le N}  \int_{\mathbf{x}} \int_{\mathbf{y}} \|\mathbf{x} - \mathbf{y}\|^2_2  \mathcal{G}(\mathbf{x}; \mathbf{x}_i, \sigma_i(\epsilon)^2 \mathbf{I}) \delta_{\mathbf{x}_i}(\mathbf{y}) d\mathbf{x}d\mathbf{y} =  \frac{1}{N} \sum_{1 \le i \le N} \mathcal{W}_2(  \mathcal{G}(\mathbf{x}_i, \sigma_i(\epsilon)^2 \mathbf{I}), \delta_{\mathbf{x}_i}  )^2  \\
			& =  \frac{1}{N}  \sum_{1 \le i \le N} \Big( \sigma_i(\epsilon)^2 D_{\mathrm{P}} + \| \mathbf{x}_i - \mathbf{x}_i \|^2_2 \Big) =  \frac{1}{N} \sum_{1 \le i \le N} \epsilon^2 \ln \Big(1 + D_{\mathrm{P}}  \min_{j \neq i} \| \mathbf{x}_i - \mathbf{x}_j \|_2^2 \Big)^2 \le \epsilon^2 D_{\mathrm{P}} (\ln (1 + \rho))^2.
		\end{aligned}
	\end{equation}
	Therefore, we get the conclusion: $\mathcal{W}_2(f_{\mathbf{X}, \sigma(\epsilon)}, f_{\mathbf{X}}) = \mathcal{O}(\epsilon \ln \rho \sqrt{D_{\mathrm{P}}})$.
	
\section{Proof: Regularity of the Function Space}
\label{appendix: proof of the regular space}

	For the first part, we only have to prove that the mixture representation $f_{\mathbf{X}, \sigma(\epsilon)}$ is always square-integrable. Because if this is the case, an element that is in the space $\mathcal{F}_{\mathrm{mix}}$ also belongs to the L2 space $\mathcal{L}^2(\mathbb{R}^{D_{\mathrm{X}}})$, indicating $\mathcal{F}_{\mathrm{mix}} \subseteq \mathcal{L}^2(\mathbb{R}^{D_{\mathrm{X}}})$. Specifically, we have
	\begin{equation}
		\begin{aligned}
			& \| f_{\mathbf{X}, \sigma(\epsilon)} \|_{\mathcal{L}^2}^2 = \int f_{\mathbf{X}, \sigma(\epsilon)}(\mathbf{y})^2 d\mathbf{y} = \int \Big( \sum_{1 \le i \le N} \frac{1}{N} \mathcal{G}(\mathbf{y}; \mathbf{x}_i, \sigma_i(\epsilon)^2 \mathbf{I})  \Big)^2 d\mathbf{y} \\
			& =  \int \Big( \frac{1}{N^2} \sum_{1 \le i, j \le N} \mathcal{G}(\mathbf{y}; \mathbf{x}_i, \sigma_i(\epsilon)^2 \mathbf{I}) \mathcal{G}(\mathbf{y}; \mathbf{x}_j, \sigma_j(\epsilon)^2 \mathbf{I})  \Big) d\mathbf{y} \\
			& =  \frac{1}{N^2} \sum_{1 \le i, j \le N} \Big( \int \mathcal{G}(\mathbf{y}; \mathbf{x}_i, \sigma_i(\epsilon)^2 \mathbf{I}) \mathcal{G}(\mathbf{y}; \mathbf{x}_j, \sigma_j(\epsilon)^2 \mathbf{I}) d\mathbf{y} \Big).
		\end{aligned}
	\end{equation}
	It is a known fact~\citep{ahrendt2005multivariate} that the Gaussian product can be reshaped as
	\begin{equation}
		\mathcal{G}(\cdot) 	\mathcal{G}(\cdot) = \mathcal{G} \Big(\mathbf{x}_i; \mathbf{x}_j, (\sigma_i(\epsilon)^2 + \sigma_j(\epsilon)^2) \mathbf{I} \Big) \mathcal{G} \Big(\mathbf{y}; \frac{\sigma_j(\epsilon)^2 \mathbf{x}_i + \sigma_i(\epsilon)^2 \mathbf{x}_j}{\sigma_i(\epsilon)^2 + \sigma_j(\epsilon)^2}, \frac{\sigma_i(\epsilon)^2\sigma_j(\epsilon)^2 }{\sigma_i(\epsilon)^2 + \sigma_j(\epsilon)^2} \mathbf{I} \Big).
	\end{equation}
	With this fact, the function form can be simplified as
	\begin{equation}
		\begin{aligned}
			& \| f_{\mathbf{X}, \sigma(\epsilon)} \|_{\mathcal{L}^2}^2  = \frac{1}{N^2} \sum_{1 \le i, j \le N} \Big( \int  \mathcal{G}(\mathbf{x}_i; \mathbf{x}_j, (\sigma_i(\epsilon)^2 + \sigma_j(\epsilon)^2) \mathbf{I}) \mathcal{G}(\mathbf{y}; \cdot) d\mathbf{y}  \Big) \\
			& =  \frac{1}{N^2} \sum_{1 \le i, j \le N} \mathcal{G}(\mathbf{x}_i; \mathbf{x}_j, (\sigma_i(\epsilon)^2 + \sigma_j(\epsilon)^2) \mathbf{I}) \Big( \int \mathcal{G}(\mathbf{y}; \cdot) d\mathbf{y}  \Big) \\
			& = \frac{1}{N^2} \sum_{1 \le i, j \le N} \mathcal{G}(\mathbf{x}_i; \mathbf{x}_j, (\sigma_i(\epsilon)^2 + \sigma_j(\epsilon)^2) \mathbf{I}),
		\end{aligned}
	\end{equation}
	which is finite because $\sigma_i(\epsilon)^2 + \sigma_j(\epsilon)^2 > 0$. Therefore, the mixture representation is square-integrable $ \| f_{\mathbf{X}, \sigma(\epsilon)} \|_{\mathcal{L}^2}^2 < \infty$, which proves the claim.
	
	For the second part, it is sufficient to show that the square integral of the delta representation $f_{\mathbf{X}}$ is infinite. Similar to the above derivation, we first decompose the squared norm $\|f_{\mathbf{X}}\|_{\mathcal{L}^2}^2$ as
	\begin{equation}
		\begin{aligned}
			& \int \Big( \sum_{1 \le i \le N} \frac{1}{N} \delta_{\mathbf{x}_i}(\mathbf{y}) \Big)^2 d\mathbf{y} = \int \Big( \frac{1}{N^2} \sum_{1 \le i, j \le N} \delta_{\mathbf{x}_i}(\mathbf{y}) \delta_{\mathbf{x}_j}(\mathbf{y}) \Big) d\mathbf{y} \\
			& =  \frac{1}{N^2}  \sum_{1 \le i, j \le N} \Big( \int \delta_{\mathbf{x}_i}(\mathbf{y}) \delta_{\mathbf{x}_j}(\mathbf{y}) d\mathbf{y} \Big) =  \frac{1}{N^2} \sum_{1 \le i, j \le N} \delta_{\mathbf{0}}(\mathbf{x}_i - \mathbf{x}_j) \\
			& = \frac{1}{N^2} \Big( N \delta_{\mathbf{0}}(\mathbf{0}) + \sum_{i \neq j} \delta_{\mathbf{0}}(\mathbf{x}_i - \mathbf{x}_j)   \Big) = \frac{1}{N} \delta_{\mathbf{0}}(\mathbf{0}) = \infty.
		\end{aligned}
	\end{equation}
	where $\mathbf{0}$ represents a vector full of zero. This result verifies the claim, and indicate that L2 space $\mathcal{L}^2(\mathbb{R}^{D_{\mathrm{X}}})$ is not able to support the probability measure $\mu_{\mathcal{F}_{\mathrm{delta}}}$.
	
\section{Warm-up Effect of Langevin Dynamics}
\label{appendix: lgvin dynamics}

	Langevin dynamics~\citep{coffey2012langevin} is a type of Markov chain Monte Carlo (MCMC)~\citep{andrieu2003introduction} that can convert any initial continuous distribution to a desired one. Specifically, one usually sets the initial distribution at time step $s = 1$ as a very simple distribution (e.g., Gaussian and uniform) that is easy to sample from, and the dynamics will incrementally update the distribution in terms of the score function (i.e., some statistical information about the desired distribution), such that it will converge to the desired one as $s \rightarrow \infty$.

	To be more rigorous, suppose that one would like to sample from a continuous distribution $\pi_{\mathrm{tgt}}$ and it is easy to sample from some distribution $\pi_{\mathrm{src}}$, then let us see the following Langevin  dynamics
	\begin{equation}
		\mathbf{z}^{(s + 1)} = \mathbf{z}^{(s)} + \beta \nabla \ln \pi_{\mathrm{tgt}}(\mathbf{z}^{(s)} ) + \sqrt{2 \beta} \mathbf{w}^{(s)}, \mathbf{z}^{(1)} \sim \pi_{\mathrm{src}},
	\end{equation}
	where $\mathbf{w}^{(s)}$ is a noise sampled from standard Gaussian $\mathcal{N}(\mathbf{0}, \mathbf{I})$ and $\{\mathbf{z}^{(s)}\}_{s \in \mathbb{Z}^+}$ forms a trajectory that records how an initial particle $\mathbf{z}^{(1)}$ evolves over time. We denote the marginal distribution of particle $\mathbf{z}^{(s)}$ at time step $s$ as $\pi_s$. Trivially, we have $\pi_1 = \pi_{\mathrm{src}}$, and importantly, the following limit
	\begin{equation}
		\lim_{s \rightarrow \infty} \pi_s = \pi_{\mathrm{tgt}}.
	\end{equation}
	holds. This type of convergence is proved to be of an exponential speed~\citep{xu2018global}, which also runs very fast in practice. Interestingly, this technique can be applied regardless of the type of initial distribution $\pi_0$. 
	
	In the framework of \textit{unordered flow}, we apply the Langevin dynamics to warm up the uniformly initialized particles $\mathbf{Y}^{(0)}$. As we can anticipate from the above guide, the empirical distribution of tuned particle set $\mathbf{Y}^{(S_{\mathrm{lgvin}})}$ should be consistent with the mixture representation $\widehat{f}_{\star, \sigma(\epsilon)}$. Therefore, while some particles might be initially located at low-density areas, they will be attracted to high-density regions by the Langevin dynamics. This warm-up trick is effective to make gradient-based search robust to \textit{noisy peaks}, such that the quality of final outcome $\widehat{\mathbf{X}}$ is less likely to be affected by the constant $\epsilon$ and neural network approximation errors.

\section{More Experiment Details}
\label{appendix: experiment details}

	For the Poisson process, we set the intensity function as an square-exponential mixture:
	\begin{equation}
		\lambda_{\mathrm{poisson}}(\mathbf{x}) =  \mu \sum_{1 \le i \le 3} w_i \exp(-\|\mathbf{x} - \mathbf{b}_i\|_2^2).
	\end{equation}
	In our proof-of-concept experiment, the parameters $w_i, \mathbf{b}_i, i \in \{1, 2, 3\}$ are set as
	\begin{equation}
		\begin{aligned}
			& \mu = 1.0, w_1 = 0.3, w_2 = 0.3, w_3 = 0.4, \\
			 & \mathbf{b}_1 = [2.51, 3.12]^{\top}, \mathbf{b}_2 = [-2.01, -1.12]^{\top}, \mathbf{b}_3 = [2.51, -2.31]^{\top}.
		\end{aligned}
	\end{equation}
	Regarding the Hawkes process, we adopt a self-exciting intensity function based on the Gaussian kernel:
	\begin{equation}
		\lambda_{hawkes}(t \mid \mathcal{H}) = \mu + \sum_{s \in \mathcal{H}} \alpha \exp(- \beta (t - s)),
	\end{equation}
	where $\mathcal{H}$ denotes the set of previously occurred events, with $\mu, \alpha, \beta$ set as $0.5, 0.5, 3.0$ in the experiment.

\end{document}